# SIM: A mapping framework for built environment auditing based on street view imagery

Huan Ning[1]*, Zhenlong Li[1], Manzhu Yu[1], Wenpeng Yin[2]

*corresponding author, Huan Ning: email: hning@psu.edu

[1] Geoinformation and Big Data Research Laboratory, Department of Geography, Pennsylvania State University, University Park, PA, United States

[2] Department of Computer Science and Engineering, Pennsylvania State University, University Park, PA, United States

**Abstract:**

Built environment auditing refers to the systematic documentation and assessment of urban and rural spaces' physical, social, and environmental characteristics, such as walkability, road conditions, and traffic lights. It is used to collect data for the evaluation of how built environments impact human behavior, health, mobility, and overall urban functionality. Traditionally, built environment audits were conducted using field surveys and manual observations, which were time-consuming and costly. The emerging street view imagery, e.g., Google Street View, has become a widely used data source for conducting built environment audits remotely. Deep learning and computer vision techniques can extract and classify objects from street images to enhance auditing productivity. Before meaningful analysis, the detected objects need to be geospatially mapped for accurate documentation. However, the mapping methods and tools based on street images are underexplored, and there are no universal frameworks or solutions yet, imposing difficulties in auditing the street objects. In this study, we introduced an open source street view mapping framework, providing three pipelines to map and measure: 1) width measurement for ground objects, such as roads; 2) 3D localization for objects with a known dimension (e.g., doors and stop signs); and 3) diameter measurements (e.g., street trees). These pipelines can help researchers, urban planners, and other professionals automatically measure and map target objects, promoting built environment auditing productivity and accuracy. Three case studies, including road width measurement, stop sign localization, and street tree diameter measurement, are provided in this paper to showcase pipeline usage.

## 1 Introduction

Built environment auditing is a basic process for evaluating urban and rural spaces' physical, social, and environmental characteristics, including walkability, road conditions, and the presence of street furniture, such as traffic lights and signage. Traditionally, such audits relied on field surveys and manual observations, which are labor-intensive and costly. The rise of street view imagery (SVI), exemplified by platforms like Google Street View, has become a widely used data source for built environment assessments. SVI offers detailed and geo-referenced visual records of streetscapes, providing a valuable data source for analyzing infrastructure, accessibility, and urban aesthetics.

Recent advancements in artificial intelligence (AI), especially deep learning (multi-layer artificial neural networks), have enabled automatic object detection from SVI, facilitating the identification of key elements such as sidewalks, trees, and street signs. However, while object detection in SVI is well-explored, the localization and measurement of detected objects remain under-investigated. Many existing studies focus on on-broad classification or integrating multiple sensor-based approaches used in autonomous driving, which do not always apply to publicly available SVI data (Zhao et al., 2024). Traditional photogrammetric methods, such as stereo vision and triangulation, require precise camera parameters that are often unavailable in SVI metadata. In addition, the geometric processing algorithms for street view images, especially panorama, are different from the other images, such as aerial and close-range; there are other specific processing needs to consider, such as metadata processing, paralleling image downloading, and coordinate conversion. However, the basic algorithm libraries for such processing are rare (He & Li, 2021), as well as the downstream mapping and measurement pipelines built upon these basic algorithms. As a result, there is a gap in the development of systematic methods for localizing and measuring street objects, hindering scalable applications in urban planning, mobility assessment, and environmental auditing.

To address these challenges, this study proposes a universal framework, Street Image Mapping, or SIM, for street image mapping and measuring built environment features using

SVI. The framework includes foundational processing algorithms and three pipelines: (1) width measurement for ground objects such as roads and sidewalks, (2) 3D localization for objects with known dimensions like stop signs, and (3) diameter measurement for features like tree trunks. By leveraging spatial data processing and geometric measurement techniques, the framework enables automated, accurate, and scalable mapping of built environment elements.

The remainder of this paper is structured as follows: Section 2 reviews related work on SVI-based built environment analysis, Section 3 details the methodology and implementation of the mapping framework, as well as case studies demonstrating its applications, and Section 4 discusses the learned lessons and future research. Finally, Section 5 concludes this study.

## 2 Related work

SVI has become a viable data source for measurement due to its geo-referenced and geo-oriented images. The panorama is the major product of SVI; it refers to the 360-degree images that allow viewers to look around as if they were standing at the camera location. Many SVI services also provide camera position and orientation metadata, allowing practitioners to compute pixel geolocation based on these parameters. Despite the absence of a universal framework or standardized toolkit for measurements using SVI, researchers have explored various methodologies. This section reviews key approaches, including bird's-eye view image, 3D reconstruction, and object localization. Although terrestrial LiDAR (Light Detection and Ranging) is another potential tool for measuring the built environment, its availability remains limited compared to SVI. Therefore, this article focuses exclusively on SVI-based methods.

2.1 Bird's eye view image

Bird's Eye View (BEV) refers to a top-down perspective of a location, typically generated from image-based or sensor-derived data. BEV representations are widely utilized in autonomous driving, urban planning, and remote sensing to analyze street layouts, pedestrian infrastructure, and traffic networks. In geospatial sciences, similar terms such as land cover maps, orthomaps, or overhead images are used interchangeably with BEV imagery; we adopt these terms interchangeably in this paper.

The conversion of vehicle-mounted camera images into land cover maps or BEV representations has been extensively studied (Can et al., 2022; Du et al., 2023; Jain et al., 2021; Schulter et al., 2018; Z. Wang et al., 2019). The core principle involves using images or depth data to reconstruct a 3D semantic environment. These reconstructions typically require highly accurate data from well-calibrated sensors such as cameras and LiDAR (Ng et al., 2020; Paek et al., 2021; Xu et al., 2022), and they rely on sophisticated data-processing algorithms and models. Deep learning models, including Generative Adversarial Networks (Goodfellow et al., 2014), have been widely employed for high-fidelity scene reconstruction. Intensive sensor data and computational resources enable the generation of high-definition maps that document the surroundings of autonomous vehicles. However, these studies predominantly focus on detecting and mapping roadway surfaces and dynamic objects, such as vehicles and pedestrians (Badue et al., 2021), rather than assessing built environment features beyond the road.

The geographic information science (GIScience) community is increasingly exploring applications of BEV imagery for environmental auditing. For instance, Ning et al. (2022) utilized depth map data from Google Street View images to convert street-level imagery into land cover maps for sidewalk width measurement. In this study, we generalize and integrate their method into our framework as the width measurement pipeline for automatic and scalable built environment auditing.

2.2   3D reconstruction using street view imagery

Mapping objects for built environment auditing using SVI aligns with image-based 3D reconstruction (Niu et al., 2024), a well-established approach in photogrammetry (Linder, 2009) that enables accurate 3D measurements from photographs. Traditional photogrammetry relies on control points with known coordinates to restore the position and orientation of images for 3D scene reconstruction. Additionally, precise camera parameters, such as lens distortion, are required to establish rigorous geometric models that define the relationship between the camera center, photo plane, and physical space.

3D reconstruction is a fundamental task in photogrammetry and computer vision, where a three-dimensional model of an environment is generated from two-dimensional images (Christodoulides et al., 2025; Luo et al., 2024). Traditional methods (Jiang et al., 2024) rely on stereo vision, multi-view geometry, and structure-from-motion (SfM) (Özyeşil et al., 2017) to infer depth and spatial relationships between detected objects. In the recent decade, researchers have explored deep learning methods to extract depth information from SVI in end-to-end manners. In reviewing the vision-based 3D occupancy prediction, Christodoulides et al. (2025) listed various models to restore road scenes for autonomous driving. Pang & Biljeck (2022)'s study demonstrated that single SVIs can be used to create coarse building 3D models using Pix2Mesh (N. Wang et al., 2018). Developed by Guo et al. (2023), the StreetSurf approach divided the SVI into three parts, close-range, distant-view, and sky, and then adopted different neural networks to restore scene surfaces. Overall, the techniques for 3D reconstruction or depth extraction are well studied and can be tailored and applied for built environment auditing. However, such practices are less investigated in the literature.

2.3    Object Localization in Street View Imagery

Mapping detected objects from SVI means to determine the objects' geolocations. The SVI 3D reconstruction results are usually centered on the camera, i.e., the obtained object distance is referenced to the camera, and the direction is related to the camera heading. The following localization requires the camera positions and orientation, which are usually provided in the SVI metadata. According to the geometric relationship between the object, pixel position, and camera location, the object can be located. Many SVI applications adopted such pipelines to locate target objects.

Tsai & Chang (2013) proposed a method utilizing Google Street View images to measure Points of Interest (POIs). By applying the forward intersection method to two overlapping street view images, users were able to determine the 3D positions of POIs with an error of approximately 6 meters, while the horizontal coordinates error is 1–2 meters. Campbell et al. (2019) employed the focal length, pixel size of the panoramic camera, and the known dimensions of traffic signs as input parameters for triangulation to calculate the distance between the camera and the traffic signs. They subsequently mapped the signs based on their azimuths and the positions of the panoramas. This method relies on precise internal camera parameters (such as focal length and pixel dimensions), which may not be obtainable for undocumented street view imagery. Based on stereoscopic photogrammetry, Eliopoulos et al. (2020) computed the tree diameter; they collected tree photos using a RealSense D435 stereo camera. Their results show a high precision of less than 2 centimeters. Although Eliopoulos et al. (2020)'s work did not use public street view images, it showcased the feasibility of diameter measurement for street trees using photogrammetry. Overall, object measurement based on SVI has been used in various applications by considering known dimensions (e.g., camera focal length) and geometric relationships. These studies did not release reproducible programs; our work attempts to generalize and incorporate these approaches into a universal framework.

## 3   Mapping pipelines

This section introduces the three mapping pipelines in the framework: width measurement for ground objects, 3D localization for objects with a known dimension, and diameter measurement. The research can choose the applicable pipeline for their target objects. The mapping and measurement processes based on street view imagery have inherent limitations compared to rigorous photogrammetry. These pipelines are built upon a foundational algorithm library, including data download, image operation, depth data operation, and coordinate conversion (Figure 1). The proposed framework is extendable based on the foundational library; pipelines built upon this library can be added.

One key limitation is the potential inaccuracy of camera location and orientation data, which may not be documented and can vary due to equipment upgrades. As a result, SVI-based measurements often incorporate complementary data to mitigate these limitations and streamline the process. For instance, depth maps can be derived from SVI metadata or estimated using monocular depth inference techniques. Known object dimensions, such as the standardized height of traffic signs, can also be integrated into localization methods to enhance accuracy. Another limitation is relatively low accuracy. SVI geometry is affected by stitching errors, as panoramas, composed of multiple images captured from different cameras, are not seamlessly aligned. These stitching artifacts complicate the restoration of precise geometric relationships between individual cameras and the photo plane, especially without access to the original camera structure and raw pre-stitched images. Despite the accuracy concerns, object localization based on SVIs can achieve practical location accuracy for urban environment auditing, typically within a range of 1 to 5 meters horizontally in optimal conditions. We demonstrate the applications of the proposed pipelines using a case study, respectively. Google Street View images were used in these three cases.

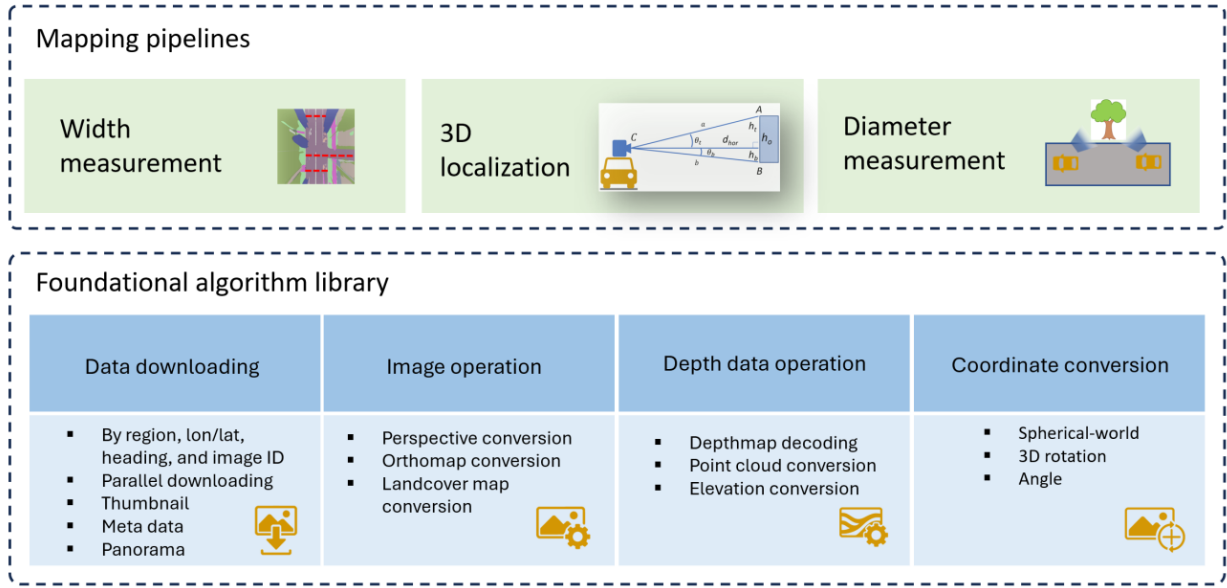

Figure 1. Mapping pipeline and foundational algorithm library

## 3.1 Width measurement for ground objects

This pipeline aims to measure the widths of ground objects along the road, such as road surfaces, sidewalks, and parking strips. The input is the land cover maps converted from street view images; the output is the measurements of the target object. We present a road width measurement case study for this pipeline.

### 3.1.1 Method

Taking the road surface as an example, Figure 2**Error! Reference source not found.** shows the width measurement method. Other ribbon objects, such as sidewalks or parking strips, can use a similar method. The main idea is to generate scanlines along the road direction and then compute the length of the intersections of the scanline and road surface. The input of this pipeline is the land cover map and street view image metadata. Below are the major steps:

Step 1: Rotate the map, making its heading direction point upward, i.e., the *y-axis* (vertical axis). This rotation makes the scanline generation mathematically easier.

Step 2: Binarize the rotated map: the target class (road surface in this case) is set to 1 and other classes is 0.

Step 3: Remove the noise pixels and gaps using a morphological operation (open-then-close).

Step 4: Generated scanlines are along the *y-axis*. The interval of scanlines can be set to a relatively small value (e.g., 0.25 m) to capture the minor changes in width. A run-length encoding (Golomb, 1966) algorithm is used to extract the starting and ending column of each target class row, i.e., the intersection of scanlines and the target class. We denote the intersection as *slice*.

Step 5: Extract the *slice* attributes, such as touching classes and cover_ratio. The former refer to the classed of two ends touched, and the latter is the ratio of the class pixel number in the minimum bounding square of the *slice* to the total pixel number of the square (Ning et al., 2022). These attributes can be used to filter *slices* in the downstream applications.

Step 6: Re-rotate the *slices* back to the original coordinates. Then, they become the width measurements of the target class.

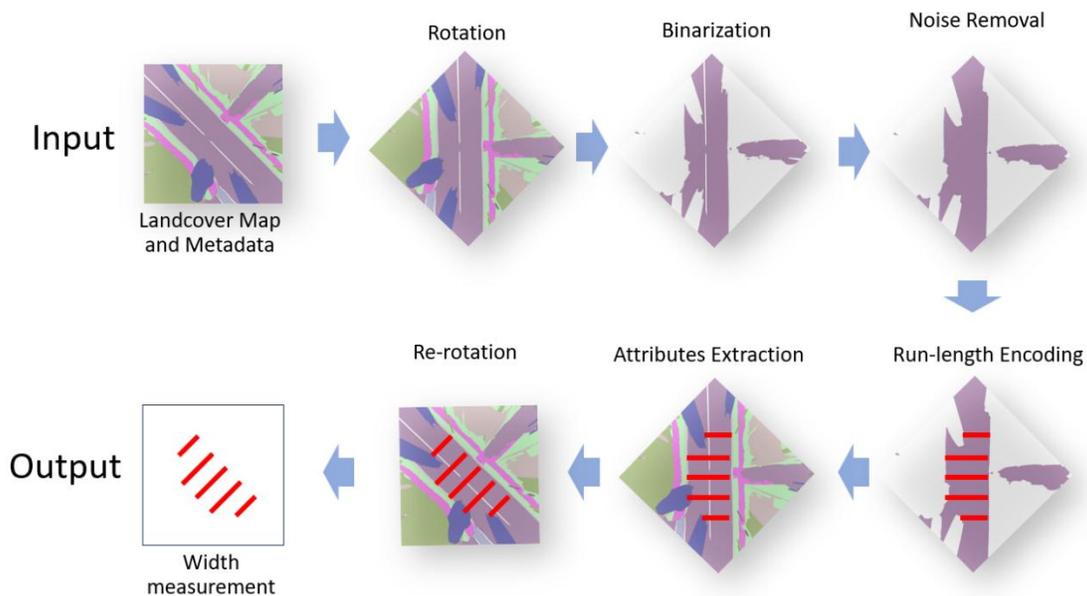

Figure 2. Workflow of the width measurement pipeline

The extracted attributes are critical for downstream applications because many of these measurements are invalid and need to be removed according to the attributes. Figure 3 shows examples of valid and invalid *slices*. Figure 3 (a) is a land cover map converted from the panorama in Figure 3 (b) using the method proposed by Ning et al. (2022). The complex urban environment may cause many invalid *slices*, such as those occluded by the vehicles, so we developed a set of algorithms to label them. For instance, *slice* $l_1$ in Figure 3(a) touches vehicles so its length does not correctly reflect the road width; it should be removed according to the touching class in the *slice* attribute. *Slice* $l_2$ extends to another road segment; therefore, it is an invalid width measurement and can be removed according to *cover_ratio*; e.g., set *cover_ratio* > 0.9, meaning that the road pixels cover the 90% area of the minimum bounding square, and this *slice* has a large chance as a valid measurement. $l_3$ is such a valid slice as it correctly represents the road width.

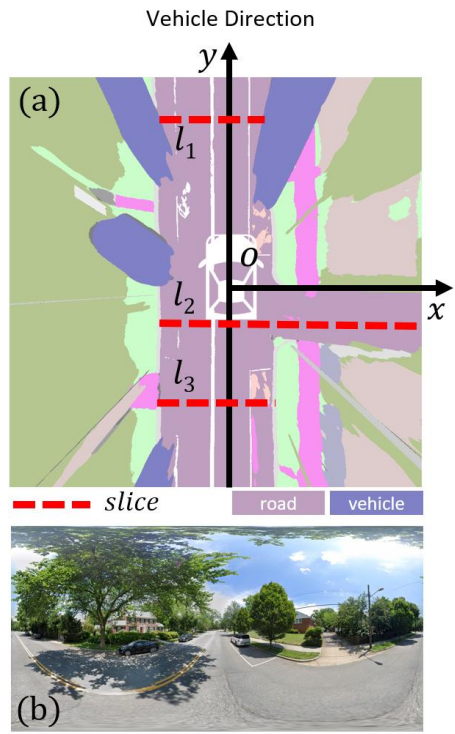

Figure 3. Road width measurement in a land cover map (a) converted from a panorama (b). The *slice* $l_1$ touches vehicles while $l_2$ extends to another road segment; both are invalid width measurements and will be ignored. Only $l_3$ will be kept in this example.

### 3.1.2 Case study: road width measurement for Wanshitong DC, US

We used Washington, D.C., US, as an study area to demonstrate the width measurement pipeline. 157,000 semantically segmented panoramas by the Seamseg model (Porzi et al., 2019) were converted to land cover maps using Ning et al. (2022)'s method. The road widths were measured by the developed pipeline, and we manually measured some road widths for accuracy assessment.

The width measurement pipeline generated 35.3 million measurements, and 8.0 million of them were valid. To assess their accuracy, we randomly and evenly selected 293 measurements, manually measured their associated actual road width, and then computed the error. The average actual width of the 293 roads is 9.76 m. The RMSE is 1.48 m (or 18.5% in relative error), the mean error is 0.85 m (10.0%), and the median is 0.56 m (6.4%). Figure 4 (a) displays the scatter plot of the ground truth and measurement from street view images, and Figure 4 (b) presents the histogram of the error. Both figures indicate that the proposed pipeline is useful and valid for measuring ground objects. Figure 5 presents some examples of width measurements. The measurements from the framework matched the actual width well.

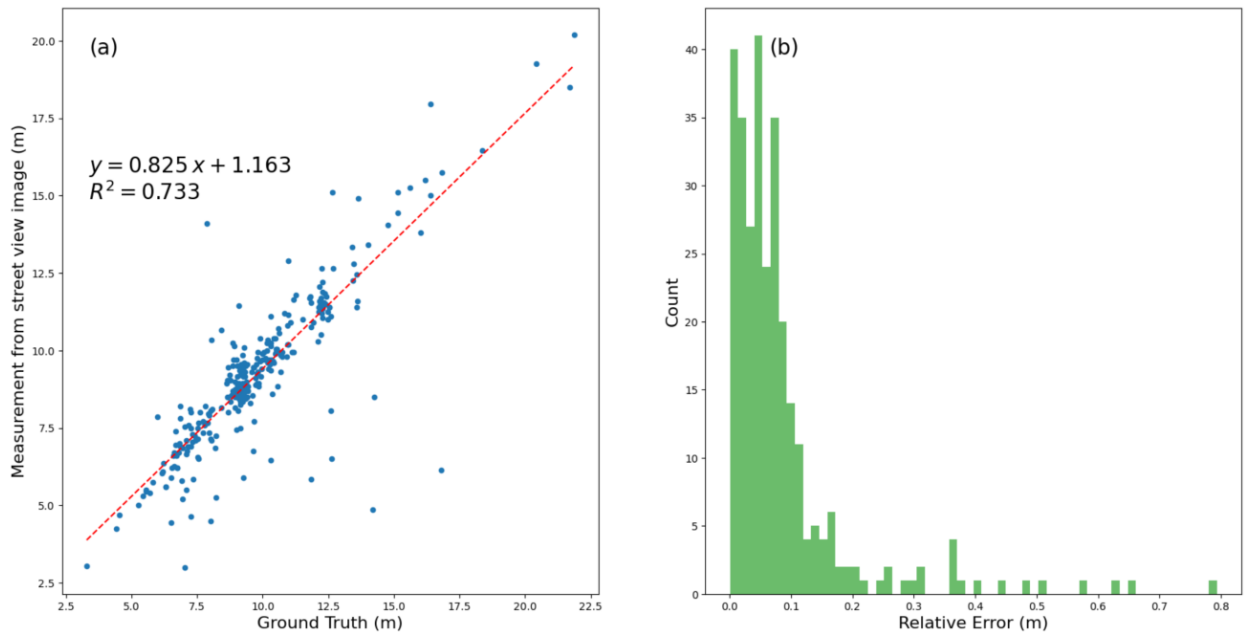

Figure 4. (a): scatter plot of road width measurement and ground truth. (b): distribution of absolute error. There are 16 outliers among the 293 measurements that have relative errors over 0.3; most measures (78%) have relative errors within 0.1.

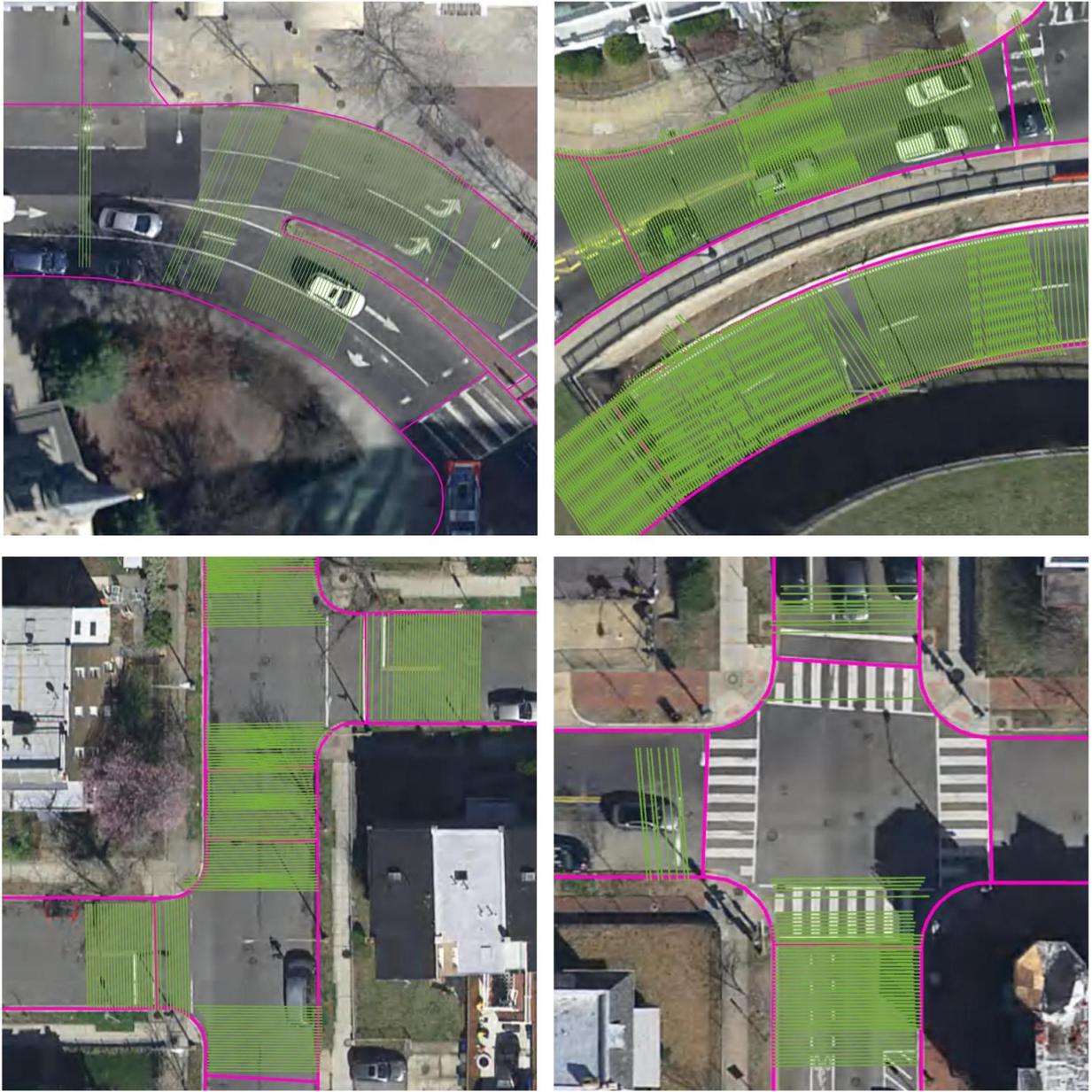

Figure 5 Road width measurement examples. The measurements from the framework matched the actual width well.

Figure 6 lists four measurements with large errors. The errors of Figure 6 (a) and Figure 6 (b) are caused by the inappropriately selected measurements, although there are other measurements that correctly present the road width. Necessary post-processing can mitigate this type of error. Figure 6 (c) shows that all measurements in this location are shorter than the ground truth, due to the misaligned concepts of the ground truth and the detected results or the inaccurate depthmap. The measurements in (d) are only one-half of the ground truth, because the other half was closed for road work when capturing the street view image.

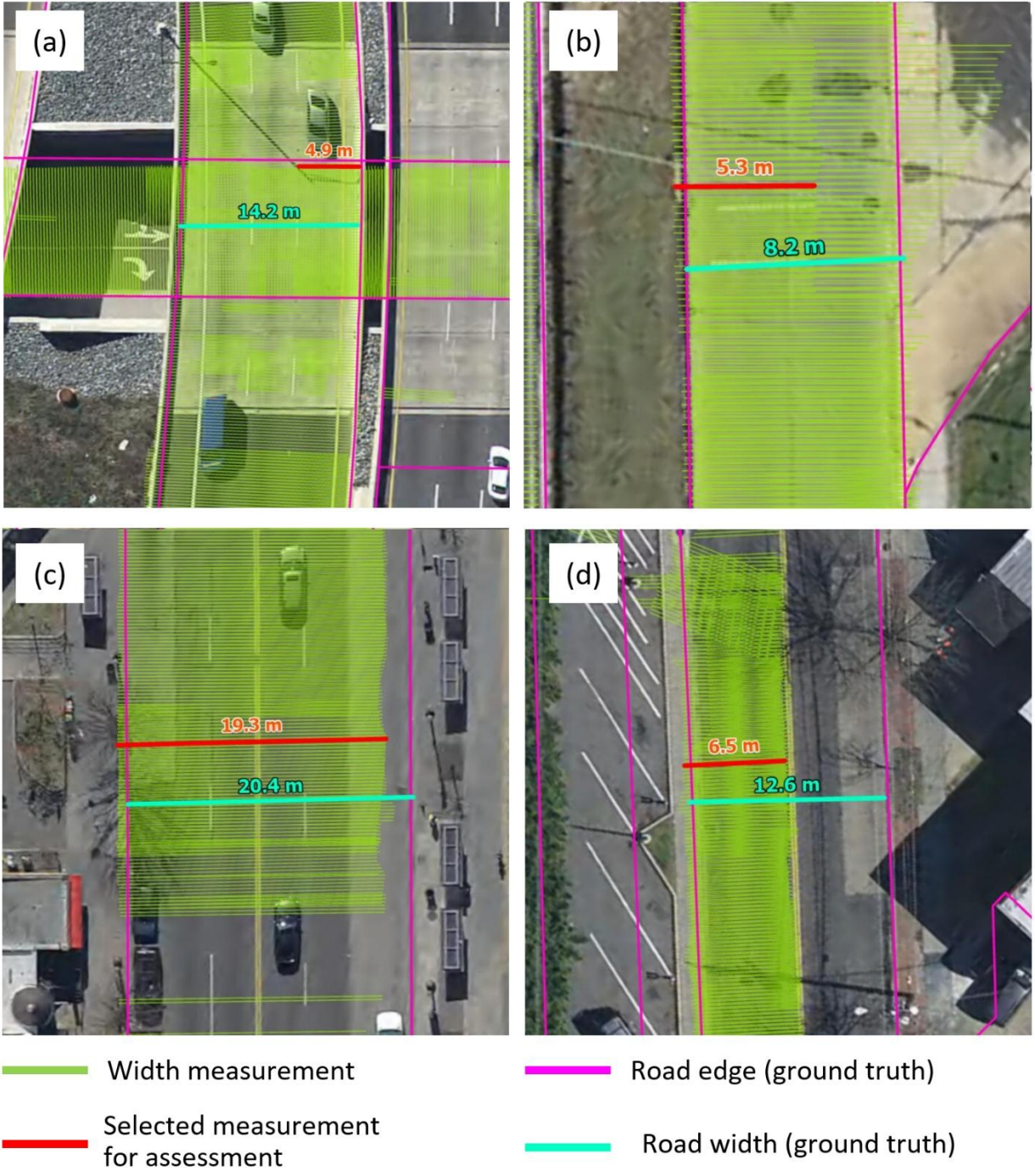

Figure 6  Samples of measurement error. The errors of (a) and (b) are caused by the inappropriately selected measurements, although there are other measurements that correctly present the road width. Necessary post-processing can mitigate this type of error. (c) shows that all measurements in this location are shorter than the ground truth, due to the misaligned concepts of the ground truth and the detected

results or the inaccurate depthmap. The measurements in (d) are only one-half of the ground truth, because the other half was closed for road work when capturing the street view image.

## 3.2 3D localization for objects with a known dimension

This pipeline utilized a known dimension of the target object for localization using single images. The main idea is to use tacheometric surveying, a quick way to measure the distance from the observer to the target object based on a known object dimension and the associated angle of its field of view (FoV) angle.

### 3.2.1 Method

This pipeline generalized the tacheometric method used in Ning et al. (2021)'s work. The object length (height or width) and the associated FoV are critical for tacheometry. Using a stop sign as an example, Figure 7 shows the workflow of the 3D localization. The input of the pipeline is the detected object bounding box, the known dimension, and image metadata (e.g., location, heading, altitude, image FoV, weight, height). First, the FoV is computed according to the bounding box and image metadata, followed by the computing of the distance from the camera to the object. The object's altitude and azimuth are also computed according to the bounding box and image metadata. Finally, the object's 3D location is calculated using the distance, altitude, and azimuth.

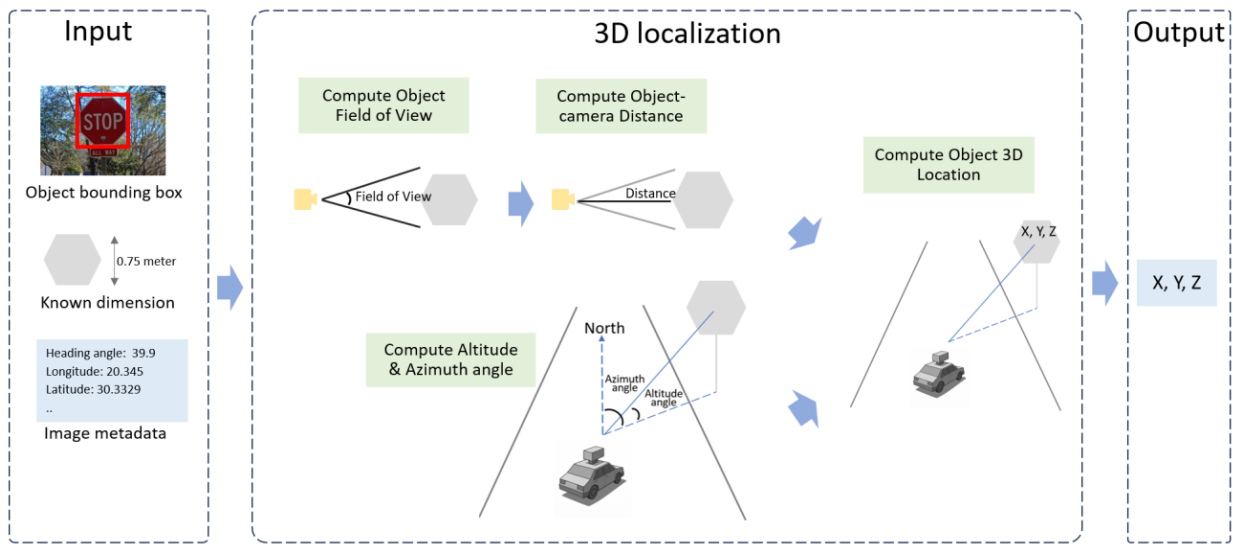

Figure 7 Workflow of 3D localization.

Specifically, we use Equation (1) and (2) to localize the object. $AB$ in Figure 8 denotes the object, and its height is known as $h_o$. Also, Figure 8 illustrates the geometric relationship between the object height ($h_o$) and its horizontal distance ($d_{hor}$) to the panorama camera $C$, where $\theta_t$ and $\theta_b$ are the altitude angle from $C$ to the object top and bottom respectively.

The area of the triangle formed by sides $a$, $b$, and $h_o$ can be calculated by $h_o \cdot d_{hor} \cdot 0.5$, or $sin(\theta_t + \theta_b) \cdot a \cdot b \cdot 0.5$, so $h_o \cdot d_{hor} = sin(\theta_t + \theta_b) \cdot a \cdot b$; by plugging in $a = d_{hor}/cos(\theta_t)$ and $b = d_{hor}/cos(\theta_b)$, we can have Equation (1) to compute $d_{hor}$. The vertical distance of the object's bottom can be obtained by Equation (2). Therefore, the 3D coordinates of the target objects originating from the panorama camera $C$ can be obtained. For further localization of the target object, the panorama camera (i.e., $C$) 's coordinate is needed, which is usually provided in its metadata from SVI services. Note that in Figure 8, Equations (1) and (2) show a special case with $A$ above $C$ and $B$ under $C$ in the vertical dimension; other conditions, including both $A$ and $B$ above or under $C$, can use similar algorithms to solve.

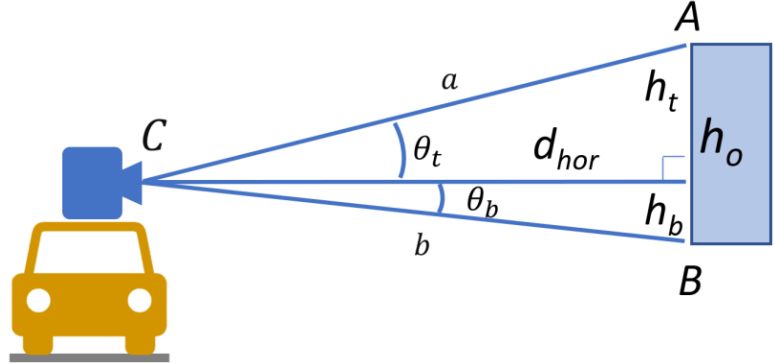

Figure 8 Image-based tacheometric surveying

$$d_{hor} = \frac{h_o \cdot cos(\theta_t) \cdot cos(\theta_b)}{sin(\theta_t + \theta_b)} \qquad (1)$$

$$h_b = tan(\theta_b) \cdot d_{hor} \qquad (2)$$

3.2.2  Case study: stop sign localization

To demonstrate the 3D localization pipeline, we took the stop sign as the target object as a study case. First, a trained YOLOv5 neural network (Ultralytics/Yolov5, 2020/2021) was used to detect the stop sign bounding box in the street view image accurately, and then the resulting bounding box position and FoV were used to calculate the vertical and horizontal coordinates. Figure 9 shows a sample of a stop sign and its detection bounding box. Trained on 642 images, YOLOv5 achieved a precision of 0.9997 and a recall of 0.9938 on a test set of 162 images.

We downloaded a traffic sign dataset of Maryland State, US. (Lloyd, 2022), containing 17,049 stop signs, as the ground truth to assess the localization error of the framework. The stop-sign height was set to 0.75 m (FHWA, 2022). This dataset did not contain vertical coordinates, so we only assessed horizontal error. We viewed their coordinates as ground truth, although accuracy is undocumented. The derived sign locations were compared with ground truths in the assessment. We downloaded 678,000 street view thumbnails for 16,897 stop signs. The street view images did not shoot the front side of every stop sign, especially those located on the side roads where the image was usually captured on the main road. Note that in this paper, a thumbnail refers a perspective image reprojected from a panorama.

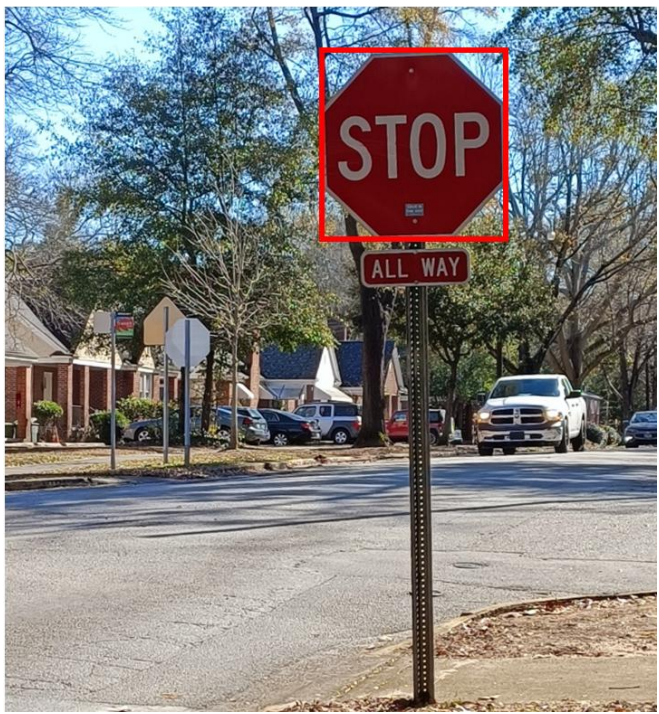

Figure 9  An example of a stop sign and its detected bounding box (red)

The 3D localization pipeline obtained 20,488 measurements for 10,423 (61.7%) stop signs; compared with the collected dataset, the horizontal RMSE (root-mean-square error) is 2.54 m, the mean error is 1.96 m, and the median error is 1.49 m. The assessments are consistent with previous studies, i.e., horizontal mean error was around 2 meters (Bruno & Roncella, 2019; Campbell et al., 201b; Krylov et al., 2018; Ning et al., 2022; Tsai & Chang, 2013). The 3D localization pipeline did not refine the location error of panoramas using external control points; therefore, the error is subject to the panorama location. Most detected stop signs (65.9%) were located close to the actual places within 2 m (see Figure 10 and Figure 11). We think this accuracy shows the practicability of this pipeline.

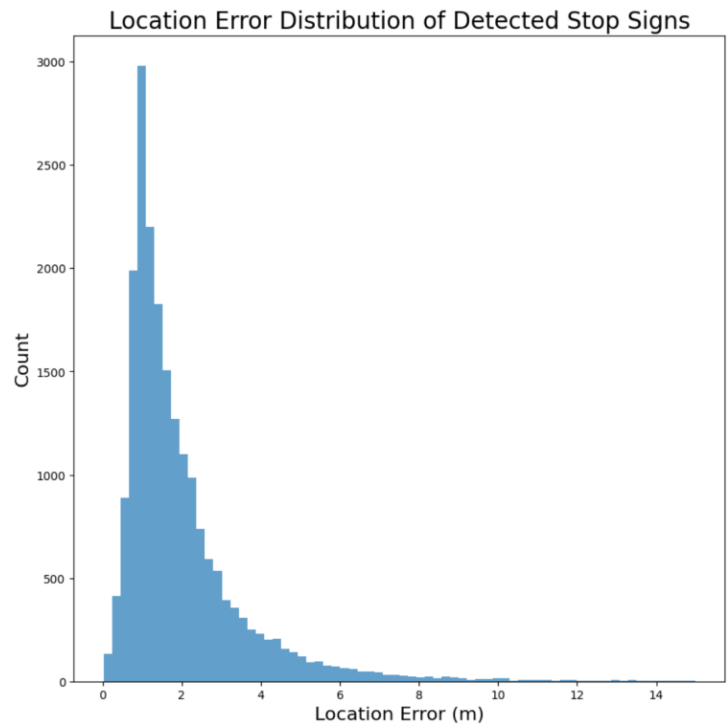

Figure 10. Loccation error distribution of detected stop signs. 65.9% of them have errors less then 2 meters.

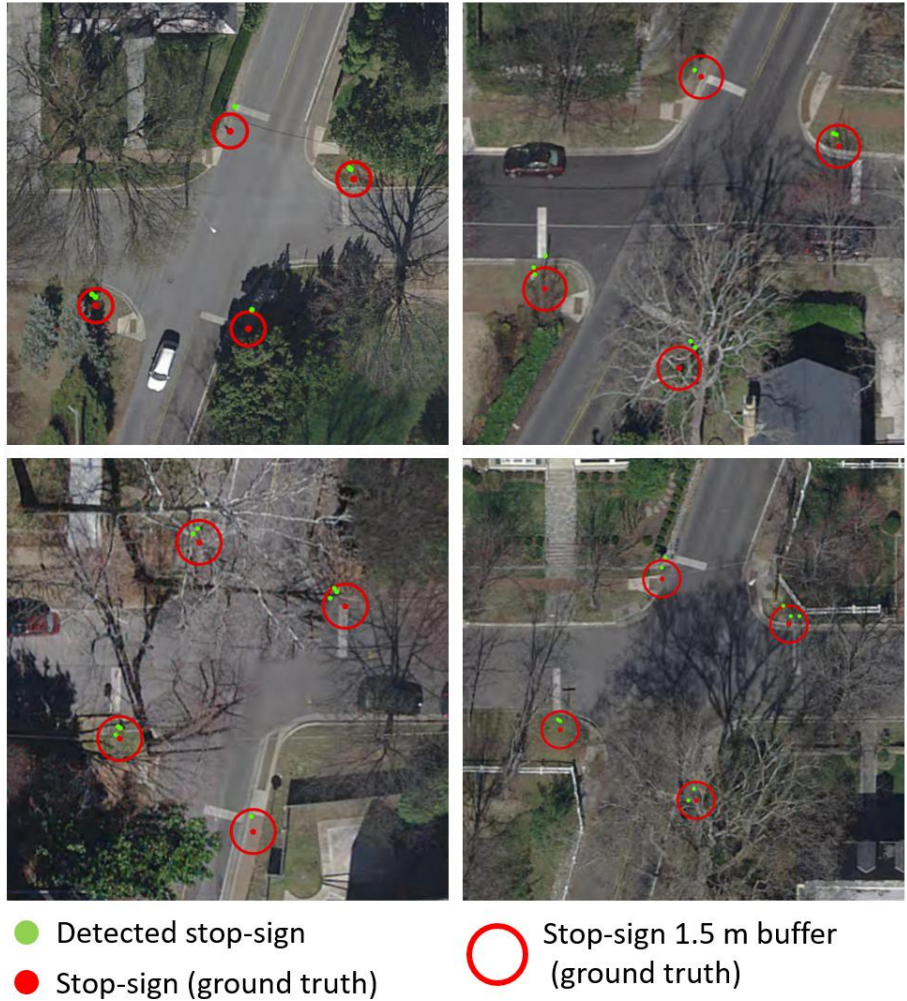

Figure 11. Stop sign localization result examples

Figure 12Figure 11 shows four localized stop signs with large reported errors. Example (a) and (b) are stop signs of parking lot exits, but they have no associated ground truths and were mistakenly matched to other stop signs, resulting in large errors. (c) shows a detected sign stop but was geolocalized to a wrong location due to its partially depicted height. (d) is an example of wrong detected stop sign: a wall was labeled as a stop sign by the model.

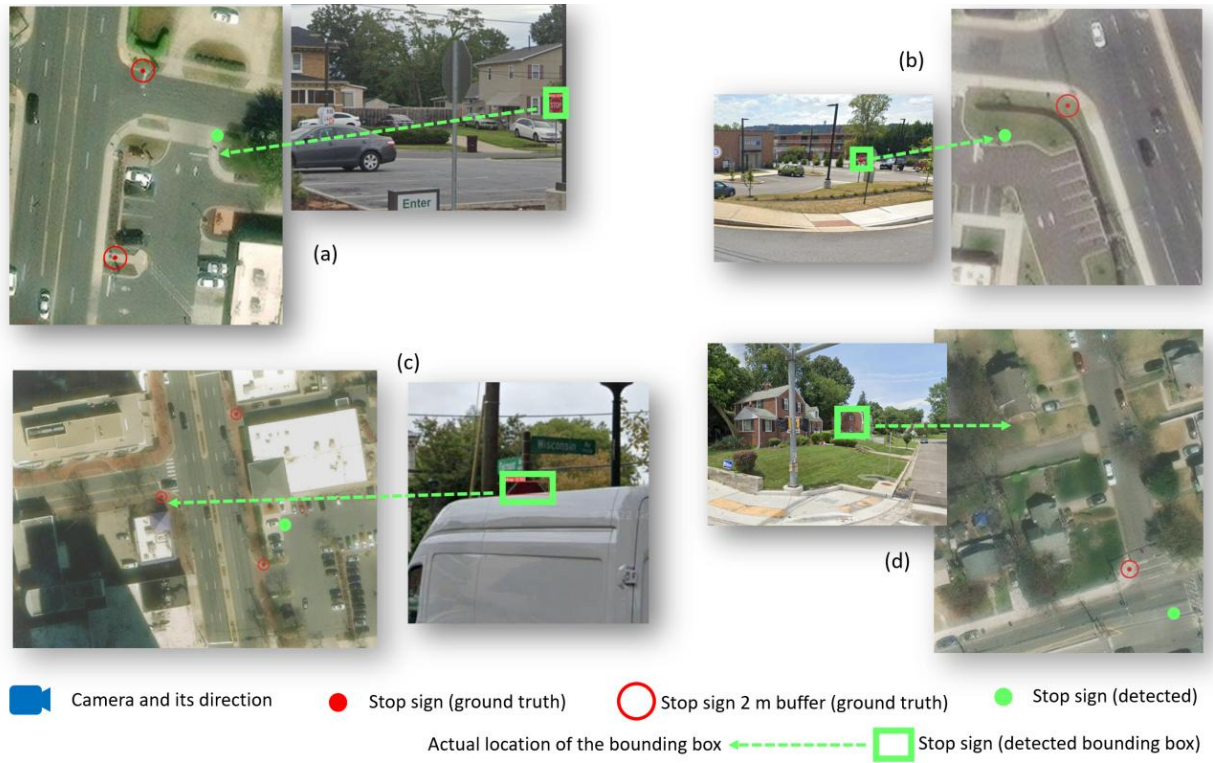

Figure 12. Examples of stop sign locations with large error. (a) and (b): the detected stop signs are correct, but large errors are mistakenly reported due to the missing records in the ground truth. (c): An partial detection results in the wrong estimated location. (d) A wall is mistakenly detected as stop sign, leading to a wrong location.

## 3.3 Diameter measurements

This pipeline uses stereo image pairs to localize target objects and measure their diameters. Theoretically, photogrammetric methods using stereo images can measure multiple dimensions, including width, height, and depth, if the measured vertices can be identified in two images. However, identifying the corresponding vertices in two images is challenging and beyond the scope of this study. Therefore, this pipeline simplifies the photogrammetric process into triangulation for diameter only, i.e., measuring a horizontal width along with a vertical object, such as a tree trunk and a pole.

### 3.3.1 Method

The diameter measurement pipeline requests the diameter locations on two images as the input and then returns the diameter measurement. The main idea is to use observations from two positions toward the same diameter to establish the geometric relationship. The unknown diameter can be solved by creating two equations based on the geometric relationship between the diameter, FoV angles, SVI orientation angle, and distances of two images.

First, the image pair should be constructed. Usually, two overlapping thumbnails from two adjacent panoramas can form a pair. Although a target object may be observed on non-adjacent images, the non-adjacent image pairs are discouraged since these observations are not necessarily on the same side, and the object may be blurred due to the far distance from the camera.

After obtaining the image pair and the images' orientations and locations, we can use triangulation to derive the object location by Equation (1) – (3) according to Figure 13 (a). According to Figure 13 (b), we can have Equation (4) to compute the tree diameter. Note that Equation (4) and Figure 13 (b) use a proximation. The accurate computation can be found in Eliopoulos et al. (2020)'s work. The proximation would not hurt the precision of the proposed method since the errors of manual measurement and the determination of breast height, in reality, are higher than the lost precision due to the proximation.

In Equations (1) – (4):

$A(x_a, y_a)$, $B(x_b, y_b)$: the horizontal location of an image pair.

$C(x_c, y_c)$: the object's horizontal location.

$\theta_a$, $\theta_b$, and $\theta_c$: the three interior angles of the triangle formed by two image's locations and the object.

$s_a$, $s_b$, and $s_c$: the three side lengths of the triangle.

$d$: object diameter.

$\theta$: the angular size of an object in the field of view.

$s_b$: the distance from the image camera to the object.

$$x_c = x_a + s_b \cdot \cos(\theta_a) \tag{1}$$

$$y_c = y_a + s_b \cdot \sin(\theta_a) \tag{2}$$

$$s_b = \frac{s_c \cdot \sin(\theta_b)}{\sin(\theta_c)} \qquad (3)$$

$$d = tan\left(\frac{\theta}{2}\right) \cdot s_b \cdot 2 \qquad (4)$$

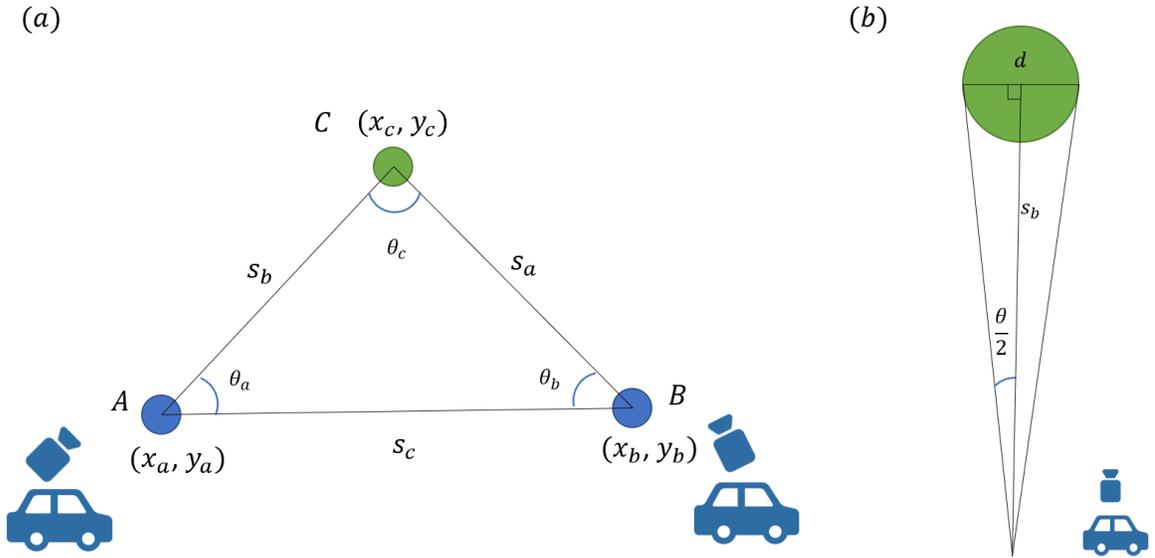

Figure 13. Localize the object (a) and compute its diameter (b)

### 3.3.2 Case study: street tree diameter measurement

We used the diameter pipeline to measure 83 street trees' diameters at South Harden St and Heyward St, Columbia, SC, USA. Manually measured diameters of these trees were obtained as ground truth for the result assessment. Before measuring diameters, the image pairs need to be formed. Figure 14 shows the workflow of the image pairing. After downloading the images, the tree trunk is segmented by a pre-trained neural network model, Grounded-SAM (Kirillov et al., 2023; Liu et al., 2024; Ren et al., 2024) (prompt: "tree trunk"). Then, the tree root and diameter measuring locations are labeled. Note that we did not adopt diameter at breast height (DBH) because it is challenging to identify the ground and the breast height (1.37 meters) ("Diameter at Breast Height," 2025) using straightforward methods; instead, we used the median width of the tree trunk as the diameter in this study. The depthmap was generated using a pre-trained monocular depth estimation model - DINO v2 (Oquab et al., 2023). Utilizing the generated depth from the camera to the diameter measuring location, the tree can be located using a single image. If the two tree locations from two neighboring images are within 3 meters, we consider these two tree locations are derived from the same tree, and these two images are a pair for this tree. Next, the trunk position in the two images can be used for triangulation.

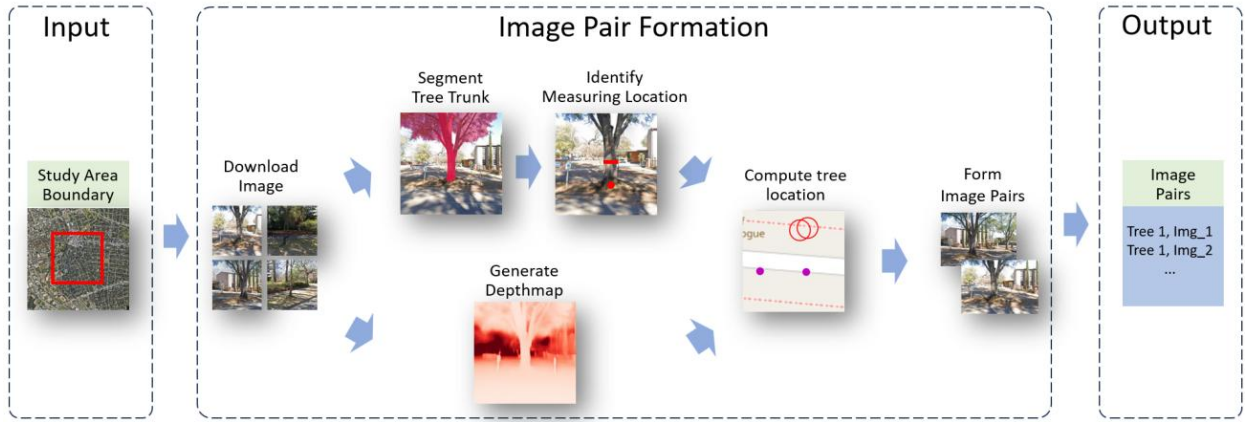

Figure 14. Workflow of image pair formation for street tree diameter measurement

The strategy of downloading image data is critical to pair formation. We designed a strategy to ensure that trees can be captured on image pairs while avoiding redundant images. First, six thumbnails (perspective images) of a panorama were collected. Referring to the mapping car's forward direction, three thumbnails toward the right 45 °, 90 °, and 135 ° respectively; similarly, another three thumbnails toward the left 5 °, 90 °, and 135 ° respectively, see Figure 15 (a). The horizontal FoV of thumbnails is 90 °, so the two adjacent thumbnails, which have a 45° heading difference on the same side, have some overlap with each other to avoid cutting trees in two images without an intact shape. We used two thumbnails from two adjacent panoramas to form an image pair for further triangulation, using the rules shown in Figure 15 (b), (c), and (d). Taking the right side as an example, three pairs can be formed: R-135° and R-90°, R-135° and R-40°, and R-135° and R-45°. The thumbnails towards the left can form another 3 image pairs. Although not implemented in this study, these image pairs can be refined after the tree direction has been determined: when obtaining the thumbnails, they can head to the tree direction and adjust the FoV to ensure the tree is clearly shown in the thumbnail.

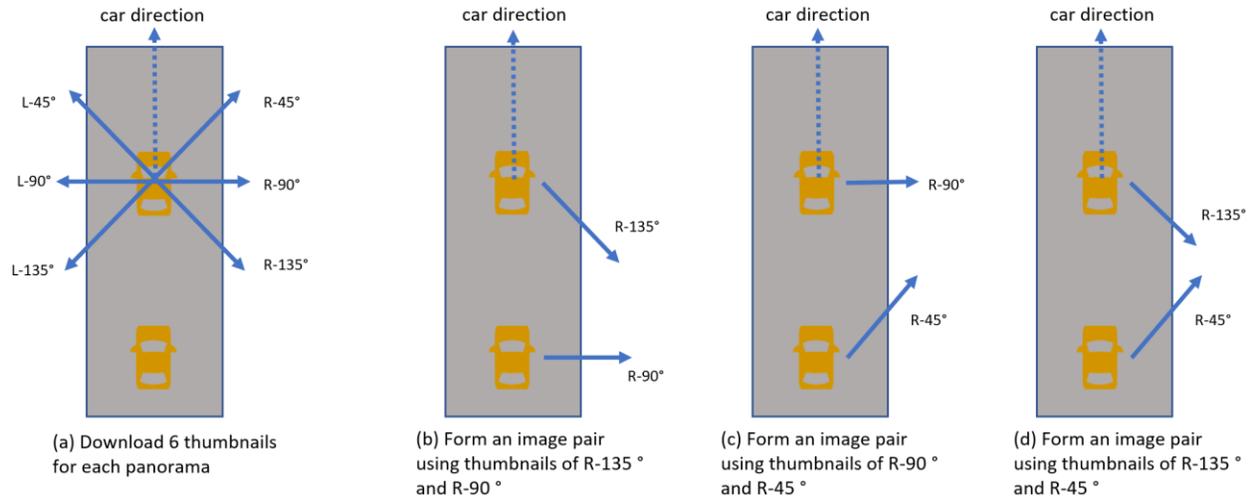

Figure 15. Strategy of street view image downloading for diameter measurement.

In the measurement results of 83 street trees, the average relative error is 7.5%, and the median is 6.8%. The measurement fill well with the ground truth, and the most relative errors (71.2%) are less than 0.1 (Figure 16). Figure 17 shows an example of the measure of street tree diameter. The top row shows the image pair from two panoramas, while the pixels are the tree trunk detected by Grounded-SAM. In this tree, the result is 0.67 meters, and the ground truth is 0.72 meters(relative error: 7%). The object detection and segmentation models (Grounded DINO and SAM) used in this study were pre-trained for general purpose without fine-tuning on tree trunk detection, so about 30% of the trees, most were small, were misdetected. Figure 18 presents three examples of measurements with large errors. It is expected that the small trunks may have large relative errors as they occupy fewer pixels, and the panorama stitching errors affect the measurement. However, for very few tree trunks, the algorithm returns large errors, but we do not find a specific reason; image geometry uncertainties may cause it.

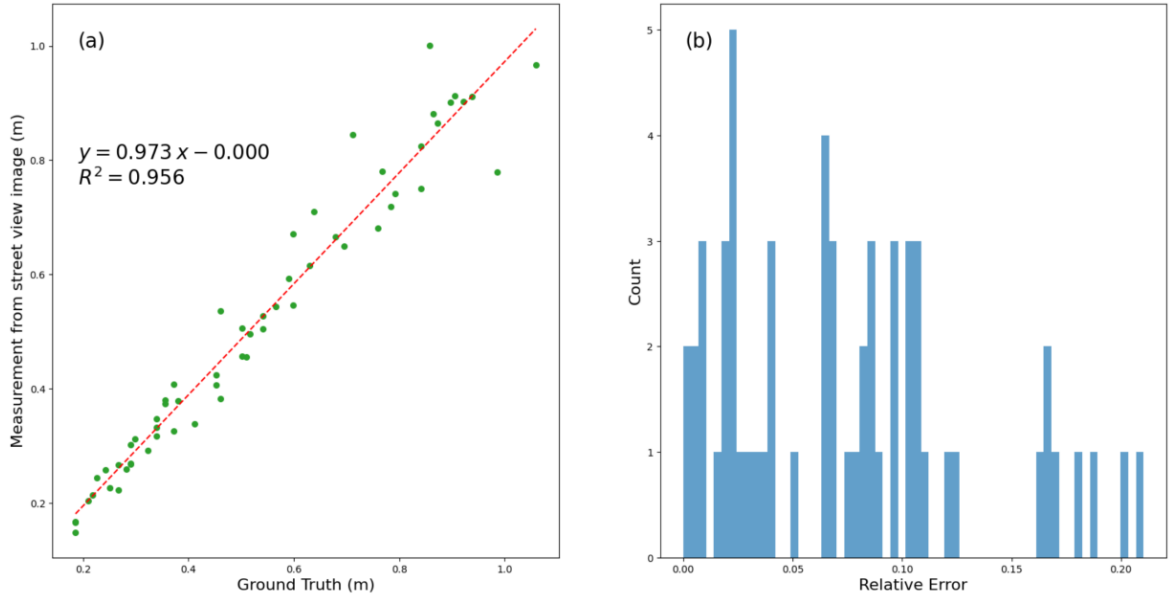

Figure 16. (a): Diameter measurements from street view image fit well with the ground truth; the association coefficient is 0.973. (b): The distribution of relative error. 71.2% measurements have relative errors less than 0.1; 8 measurements have relative errors more than 0.15.

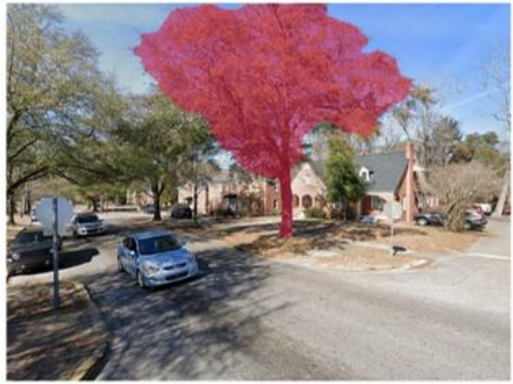
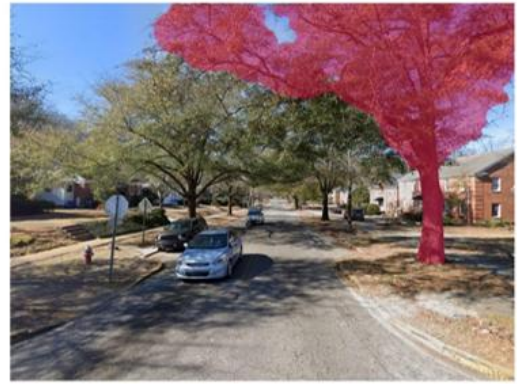

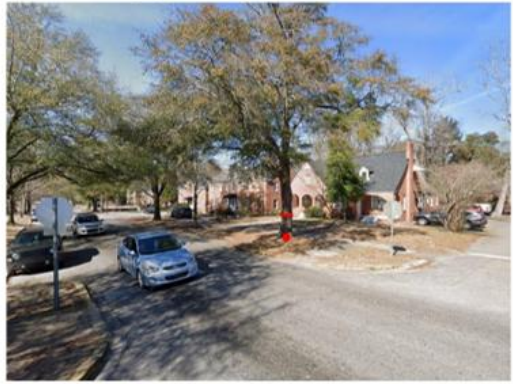
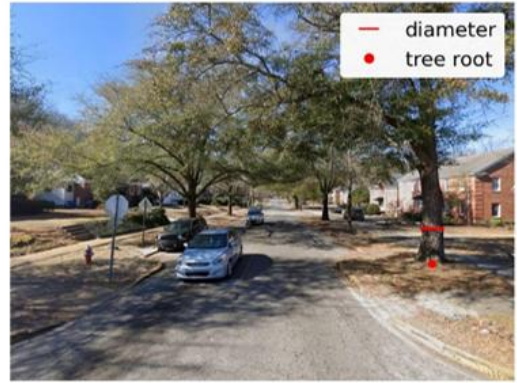

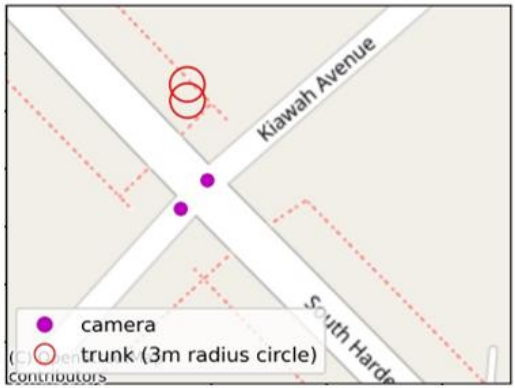
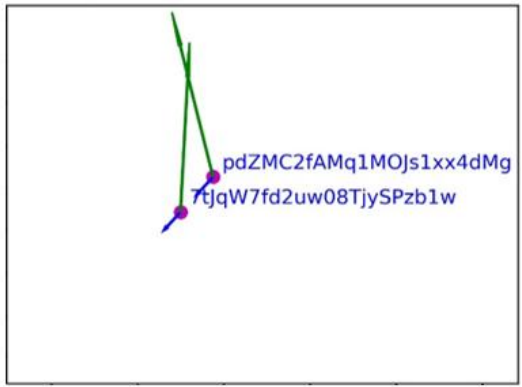

Figure 17. An example of diameter measurement: the result is 0.67 meters, and the ground truth is 0.72 meters (relative error: 7%). Top row: the red pixels are the detected tree trunks. Middle row: the red lines are the measured position for the diameter.. Bottom row: the left shows the locations of the camera and trunks (derived using monocular depth estimates); the right demonstrates trunk location from the triangulation result of the diameter measurement pipeline, the trunk location is the intersection of the two

light rays from the camera (green arrow), the blue arrows are the panoramas' orientations; and the annotations are the panorama ID.

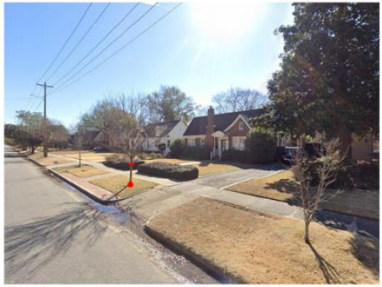 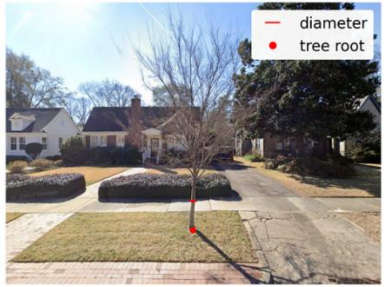

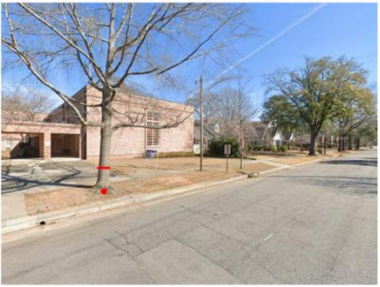 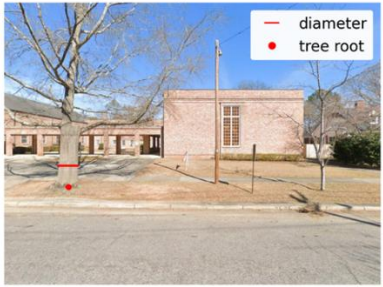

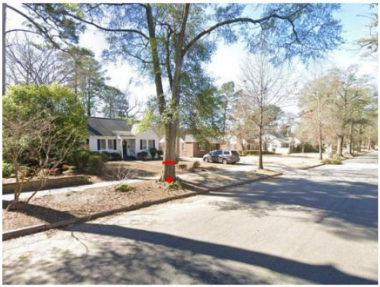 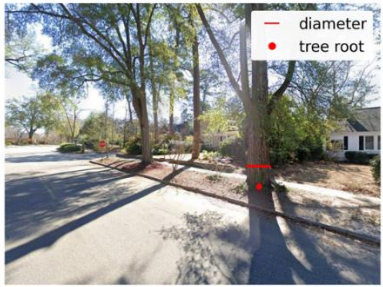

Figure 18. Causes of large errors. (a): the tree trunk is small, leading to an inaccurate boundary in the segmentation results. (b): the image has a stitching error, making the trunk larger than actual. (c): the algorithm returns a large error for this tree, but we did not find the reason; it may be caused by other image geometry uncertainty. Note that the mean diameter in this figure is computed only from the two images shown.

## 4  Discussion and future research

### 4.1  Obtaining the known length challenges the object mapping based on street view images

Obtaining the known length for triangulation is the most challenging task in street view mapping. Triangulation is a basic principle in mapping and surveying: it uses the known side lengths and angles in a triangle to compute the unknown side lengths and angles using trigonometric functions. Therefore, the known angles and lengths are required for mapping and measurement based on street view images. The angles are relatively easy to obtain from street view images since they have rigorous geometric relationships that seamlessly align with the physical street scene. The altitude and azimuth angle of the target object can be converted from the pixel position (i.e., row and column number), and the FoV angle can be derived from the range of altitude and azimuth angles. The tricky part is to get the known lengths. Regarding mapping based on a single image, external depth information is needed, such as monocular depth estimation or depth map from metadata. The accuracy of monocular depth estimation is promising; in the case study of street tree diameter measurement, the trunk location derived from the estimated depth usually has a precision of 3 meters, which is a practical precision for street tree mapping. Other parts of the depth map, such as the ground, need further inspection. The depth data from street view image metadata is an off-the-shelf data source (Ning et al., 2022) for ground object measurement, although it has been heavily simplified. Another method is using prior knowledge, such as using the known dimension of the target object (e.g., traffic signs and front doors). When mapping on stereo images, the distance between the camera locations can be used as known lengths. This method is relatively complex because it involves two images and needs to identify the same object on both images, which is complicated. Overall, obtaining sufficient and accurate known length for mapping based on publicly accessible street view images is still challenging as the applicable objects and methods are limited. We advocate that practitioners and researchers explore more objects and methods.

### 4.2  Geometric information of the target object deserves further investigation

Contemporary object detection techniques have been widely used in various domains, such as urban environment auditing. Researchers can use pre-trained or customized models to detect target objects, obtaining the bounding boxes, contours, or pixels. However, their geometric information is under-explored. We list some examples: 1) The dimensions, such as width and height, may be subject to specific criteria and have known lengths, including traffic lights, parking spots, parking meters, crosswalks, road drains, streetlights, utility holes, post boxes, and trash bins. 2) An object's actual size is identical, so the variances among stereo images can be used to determine its location. 3) The object's position on the panorama indicates its altitude and azimuth angles. 4) The length of the house's front wall can be obtained from the external building footprint. The aforementioned geometric information is investigated less in the literature. In this study, we introduced the attempts to consider the geometric information of the detected objects to benefit urban environment auditing, and encourage the research community to explore applications of the geometric information.

4.3  Limitations and future research

This study provided a framework to map and measure objects from street view images; however, the toolkit still has some limitations. First, the need for sophisticated algorithms, such as image pairing and results assessment, remains profound, but they have not been generalized and incorporated into the proposed framework. These functions should be added in future research. Second, the post-processing functions are also needed. For example, the result assessment module is to be added. One challenge is that the pre-processing and post-processing functions are complicated, requiring the consideration of the various corner cases, such as removing the unqualified measurements. After applying the mapping pipeline to more cases, the pre-processing and post-processing functions can be generalized and implemented ssto fit most scenarios.

Future research can incorporate open vocabulary object detection models, such as Grounded-SAM (Kirillov et al., 2023; Liu et al., 2024; Ren et al., 2024), into the toolkit. We expect that the users can detect and map the target object using text prompts without considering model training and image downloading. Per our experience, one issue is that pre-trained models may have a weak capability to detect street objects, such as driveways and crosswalks. Fine-tuning or customized training is needed before introducing them into the toolkit because the corrected detection is the premise of mapping and measuring.

## 5  Conclusion

This paper introduces SIM, a mapping framework for built environment auditing based on street view imagery. The framework integrates three pipelines, including width measurement for ground objects, 3D localization of objects with known dimensions, and diameter measurement for features like tree trunks. SIM automates labor-intensive traditional methods, significantly enhancing the efficiency of built environment auditing. We use three case studies to demonstrate the usage of these three pipelines, respectively. The road width measurement case shows a mean error of 0.85 meters (10%), which is fair for urban environment auditing. Using the 3D localization pipeline, the stop sign localization case demonstrates that the horizontal error is about 2 meters, consistent with previous research. The diameter measurement pipeline is showcased by measuring the street trees. The error is about 7.5% (4 cm). These case studies show that the proposed methods of automatic mapping and measurement for urban environment auditing are promising. With worldwide coverage of the street view image, SIM's automatic pipeline empowers urban environment auditing regardless of geographic extent. However, the study also acknowledges limitations, such as the need for more pre-processing functions and post-processing functions. Regarding future work, open-vocabulary object detection models can be incorporated into the frameworks.

**Data and code availability statement:**

Once the paper is accepted, the Python code can be found at: https://github.com/gladcolor/street_image_mapping

Jiang, S., You, K., Li, Y., Weng, D., & Chen, W. (2024). 3D reconstruction of spherical images:
A review of techniques, applications, and prospects. *Geo-Spatial Information Science*,
*27*(6), 1959–1988. https://doi.org/10.1080/10095020.2024.2313328

Kirillov, A., Mintun, E., Ravi, N., Mao, H., Rolland, C., Gustafson, L., Xiao, T., Whitehead, S.,
Berg, A. C., Lo, W.-Y., Dollár, P., & Girshick, R. (2023). *Segment Anything* (No.
arXiv:2304.02643). arXiv. https://doi.org/10.48550/arXiv.2304.02643

Krylov, V. A., Kenny, E., & Dahyot, R. (2018). Automatic Discovery and Geotagging of Objects
from Street View Imagery. *Remote Sensing*, *10*(5), Article 5.
https://doi.org/10.3390/rs10050661

Linder, W. (2009). *Digital Photogrammetry*. Springer Berlin Heidelberg.
https://doi.org/10.1007/978-3-540-92725-9

Liu, S., Zeng, Z., Ren, T., Li, F., Zhang, H., Yang, J., Jiang, Q., Li, C., Yang, J., Su, H., Zhu, J.,
& Zhang, L. (2024). *Grounding DINO: Marrying DINO with Grounded Pre-Training for
Open-Set Object Detection* (No. arXiv:2303.05499). arXiv.
https://doi.org/10.48550/arXiv.2303.05499

Lloyd, J. (2022, August 23). *MDOT GIS Data*. https://data-
maryland.opendata.arcgis.com/pages/mdot-data

Luo, H., Zhang, J., Liu, X., Zhang, L., & Liu, J. (2024). Large-Scale 3D Reconstruction from
Multi-View Imagery: A Comprehensive Review. *Remote Sensing*, *16*(5), Article 5.
https://doi.org/10.3390/rs16050773

Pang, H. E., & Biljecki, F. (2022). 3D building reconstruction from single street view images using deep learning. *International Journal of Applied Earth Observation and Geoinformation*, *112*, 102859. https://doi.org/10.1016/j.jag.2022.102859

Porzi, L., Bulo, S. R., Colovic, A., & Kontschieder, P. (2019). Seamless scene segmentation. *Proceedings of the IEEE Conference on Computer Vision and Pattern Recognition*, 8277–8286.

Ren, T., Liu, S., Zeng, A., Lin, J., Li, K., Cao, H., Chen, J., Huang, X., Chen, Y., Yan, F., Zeng, Z., Zhang, H., Li, F., Yang, J., Li, H., Jiang, Q., & Zhang, L. (2024). *Grounded SAM: Assembling Open-World Models for Diverse Visual Tasks* (No. arXiv:2401.14159). arXiv. https://doi.org/10.48550/arXiv.2401.14159

Schulter, S., Zhai, M., Jacobs, N., & Chandraker, M. (2018). Learning to look around objects for top-view representations of outdoor scenes. *Proceedings of the European Conference on Computer Vision (ECCV)*, 787–802.

Tsai, V. J. D., & Chang, C.-T. (2013). Three-dimensional positioning from Google street view panoramas. *IET Image Processing*, *7*(3), 229–239. https://doi.org/10.1049/iet-ipr.2012.0323

*Ultralytics/yolov5*. (2021). [Python]. Ultralytics. https://github.com/ultralytics/yolov5 (Original work published 2020)

Wang, N., Zhang, Y., Li, Z., Fu, Y., Liu, W., & Jiang, Y.-G. (2018). *Pixel2Mesh: Generating 3D Mesh Models from Single RGB Images* (No. arXiv:1804.01654). arXiv. https://doi.org/10.48550/arXiv.1804.01654
34